\pdfoutput=1

\documentclass[11pt]{article}

\usepackage{EMNLP2023}

\usepackage{times}
\usepackage{latexsym}

\usepackage{graphicx}
\usepackage{pgfplots} 

\usepackage[T1]{fontenc}

\usepackage[utf8]{inputenc}

\usepackage{microtype}

\usepackage{inconsolata}
\usepackage{booktabs} 

\title{Revealing Trends in Datasets from the 2022 ACL and EMNLP Conferences}

\author{Jesse Atuhurra, Hidetaka Kamigaito  \\
         Information Science Division\\ Nara Institute of Science and Technology, Ikoma, Japan. \\
        \texttt{\{atuhurra.jesse.ag2, kamigaito.h\}@naist.ac.jp } \\}

\begin{document}  
\maketitle
\begin{abstract}
Natural language processing (NLP) has grown significantly since the advent of the Transformer architecture. Transformers have given birth to pre-trained large language models (PLMs). There has been tremendous improvement in the performance of NLP systems across several tasks. NLP systems are on par or, in some cases, better than humans at accomplishing specific tasks. However, it remains the norm that \emph{better quality datasets at the time of pretraining enable PLMs to achieve better performance, regardless of the task.} The need to have quality datasets has prompted NLP researchers to continue creating new datasets to satisfy particular needs. For example, the two top NLP conferences, ACL and EMNLP, accepted ninety-two papers in 2022, introducing new datasets. This work aims to uncover the trends and insights mined within these datasets. Moreover, we provide valuable suggestions to researchers interested in curating datasets in the future.  
\end{abstract}

\section{Introduction}
\label{sec: Introduction}
Natural language processing (NLP) has grown significantly since the advent of the Transformer architecture. Transformers have given birth to a new pre-trained of large language models (PLMs). 

There has been tremendous improvement in the performance of NLP systems across several tasks. NLP systems are on par or, in some cases, better than humans at accomplishing specific tasks. 

\begin{figure}[t!]
\centering
\includegraphics[width=0.99\columnwidth]{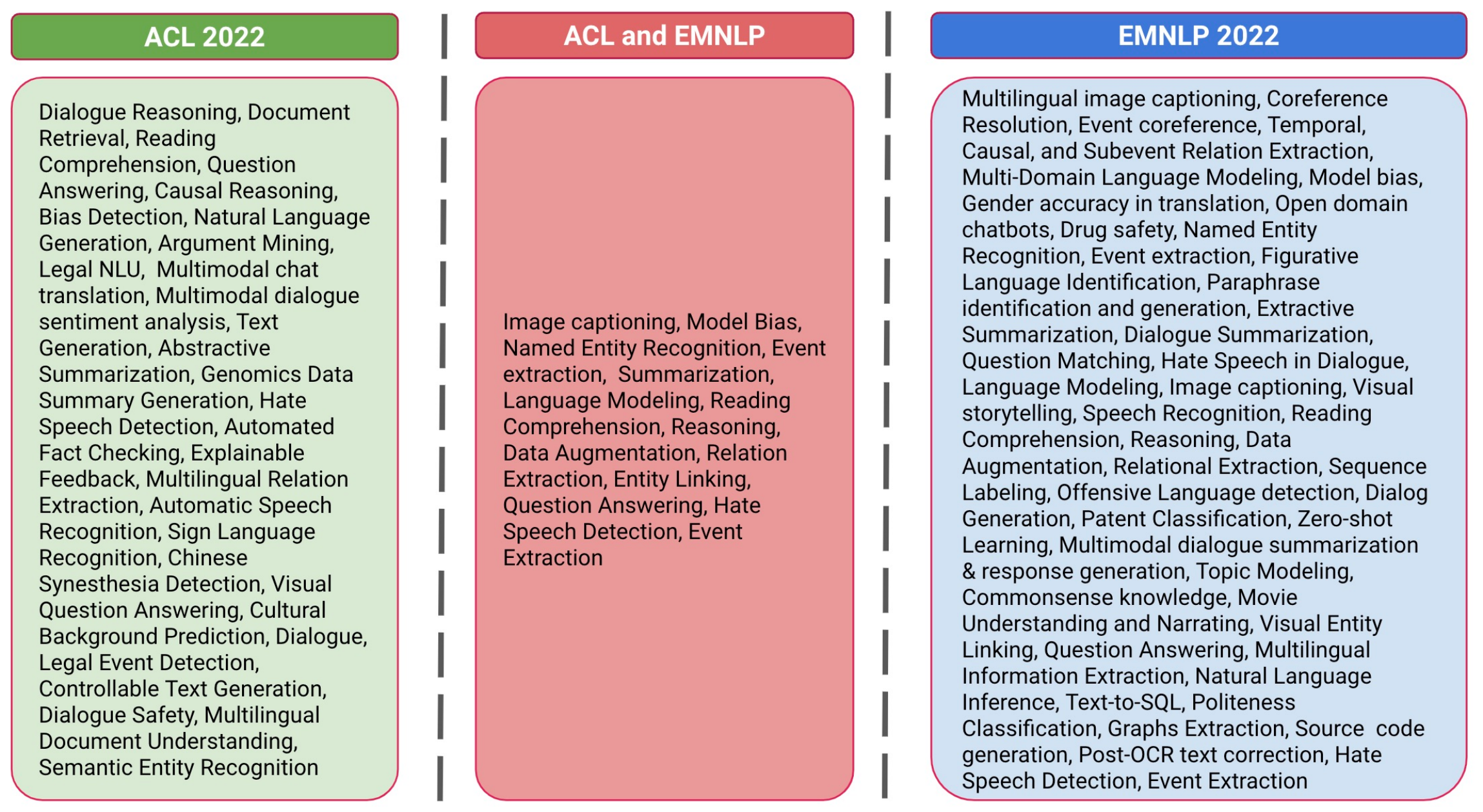}
\caption{An overview of major NLP tasks covered in datasets published in ACL and EMNLP in 2022.}
\label{fig:Intro_section}
\end{figure}

However, it remains the norm that \emph{better quality datasets at the time of training enable PLMs to perform better, regardless of the task.} The need to have quality datasets has prompted NLP researchers to continue creating new datasets to satisfy particular needs. For example, the two top NLP conferences, ACL and EMNLP, accepted ninety-two papers in 2022, introducing new datasets. This work aims to uncover the trends and insights mined within these datasets. Figure~\ref{fig:Intro_section} shows the main NLP tasks for which authors created new datasets, which were published at both EMNLP and ACL in the year 2022. The red box in Figure~\ref{fig:Intro_section} indicates the tasks in both ACL and EMNLP datasets.

We draw inspiration from \cite{gollapalli-li-2015-emnlp}, who previously studied the significant themes across NLP conferences from 1996 to 2014. The more recent work to examine the disparate sources utilized to construct new datasets is the study by \cite{yu-etal-2022-beyond}. We aim to reveal insights such as significant author affiliations in creating datasets for training NLP models. 

Lastly, we provide helpful suggestions to researchers interested in curating datasets in the future.  
\begin{figure*}[t!]
\centering
\includegraphics[width=0.95\textwidth]{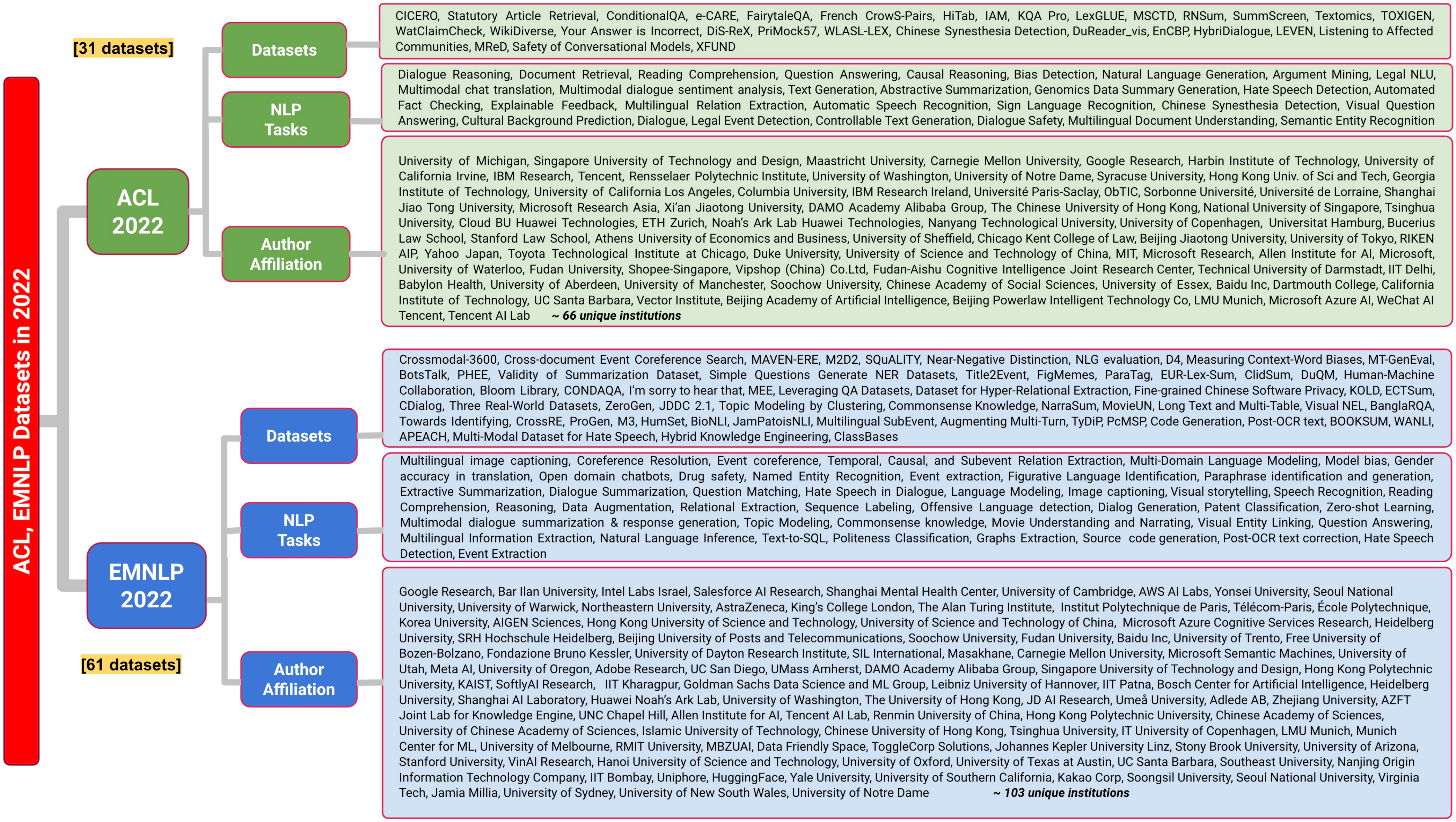}
\caption{Datasets, major NLP tasks, and author-affiliation for datasets published at ACL and EMNLP in 2022.}
\label{fig:all_DATA_2022}
\end{figure*}

\section{Scope of this Study}
\label{sec: Scope_study}
This paper's analysis and findings are based only on documents accepted at the two top-tier NLP conferences, ACL2022 and EMNLP2022. These conferences are a good measure of contemporary NLP needs; thus, the work published in these two conferences best reflects the trajectory in which NLP research is headed. \footnote{The LREC conference is a significant collection of datasets, but we aim at conferences considered top tier in NLP.} 
\paragraph{Article Retrieval} To get the relevant papers during this study, we used the keywords \emph{dataset} and \emph{corpus}. Initially, we obtained 130 papers. We then verified that each article introduced a new dataset or corpus. After this inspection, only 92 papers remained. The resulting papers were scrutinized to extract the following attributes: \emph{name of dataset, dataset size, NLP topic, primary baselines, multilingualism, multimodality, and author-affiliation}. These attributes were chosen to extract as much relevant information as possible, the kind of information that would permit us to understand the recent trend in curating new NLP datasets. 

\section{NLP Tasks Coverage in Datasets}
\label{sec: NLP_tasks_Coverage}
First, we set out to categorize the significant tasks represented in the datasets to provide a candid analysis of the datasets. Figure~\ref{fig:Intro_section} shows the significant topics that are covered by the datasets published at ACL—in the \textit{green box}; and topics covered by datasets published at EMNLP—in the \textit{blue box}. Topics common to ACL and EMNLP are shown in the middle—in the \textit{red box}.

In addition, from both Table~\ref{Table: NLP_topics_ACL} and Table~\ref{Table: NLP_topics_EMNLP}, it is evident that \emph{text generation, text summarization, text or token classification, information extraction, natural language understanding, question answering} are the focal point for dataset creation for NLP researchers. 
\begin{table}[t!]
\centering
\scriptsize 
\begin{tabular}{  p{6.5cm}   }
\toprule
\hline 
{\bf  \centerline{Major topics in ACL 2022}   }  \\ 
 \hline
Dialogue Reasoning, Document Retrieval, Reading Comprehension, Question Answering, Causal Reasoning, Bias Detection, Natural Language Generation, Argument Mining, Legal NLU,  Multimodal chat translation, Multimodal dialogue sentiment analysis, Text Generation, Abstractive Summarization, Genomics Data Summary Generation, Hate Speech Detection, Automated Fact Checking, Explainable Feedback, Multilingual Relation Extraction, Automatic Speech Recognition, Sign Language Recognition, Chinese Synesthesia Detection, Visual Question Answering, Cultural Background Prediction, Dialogue, Legal Event Detection, Controllable Text Generation, Dialogue Safety, Multilingual Document Understanding, Semantic Entity Recognition \\
\hline 
\bottomrule
\end{tabular}
\caption{\label{Table: NLP_topics_ACL} The NLP topics in datasets at ACL 2022.}
\end{table}
\begin{table}[t!]
\centering
\scriptsize 
\begin{tabular}{  p{6.5cm}   }
\toprule
\hline 
{\bf  \centerline{Major topics in EMNLP 2022}   }  \\ 
 \hline
Multilingual image captioning, Coreference Resolution, Event coreference, Temporal, Causal, and Subevent Relation Extraction, Multi-Domain Language Modeling, Model bias, Gender accuracy in translation, Open domain chatbots, Drug safety, Named Entity Recognition, Event extraction, Figurative Language Identification, Paraphrase identification and generation, Extractive Summarization, Dialogue Summarization, Question Matching, Hate Speech in Dialogue, Language Modeling, Image captioning, Visual storytelling, Speech Recognition, Reading Comprehension, Reasoning, Data Augmentation, Relational Extraction, Sequence Labeling, Offensive Language detection, Dialog Generation, Patent Classification, Zero-shot Learning, Multimodal dialogue summarization, response generation, Topic Modeling, Commonsense knowledge, Movie Understanding and Narrating, Visual Entity Linking, Question Answering, Multilingual Information Extraction, Natural Language Inference, Text-to-SQL, Politeness Classification, Graphs Extraction, Source  code generation, Post-OCR text correction, Hate Speech Detection, Event Extraction \\
\hline 
\bottomrule
\end{tabular}
\caption{\label{Table: NLP_topics_EMNLP} The NLP topics in datasets at EMNLP 2022.}
\end{table}

Table~\ref{Table: NLP_topics_ACL} indicates the list of the significant NLP topics tasks represented in the papers that introduced datasets and were published at ACL in 2022. Similarly, the NLP topics covered in the datasets published at EMNLP 2022 are shown in Table~\ref{Table: NLP_topics_EMNLP}.

\section{Dataset size}
\label{sec: Dataset_size}
The number of sentences in datasets published at ACL and EMNLP is essential to understanding the number of sentences required to build an NLP system with satisfactory performance. 
\begin{table}[!t]
\centering
\footnotesize 
\begin{tabular}{  p{3cm} | p{2cm} }
\toprule
\hline 
{\bf Data size (\# instances) }  &  {\bf Dataset Count   }  \\ 
\hline
$\le$ 500 & 3 \\
\hline
500 - 1,000 & 2 \\
\hline
1,000 - 1,500 & 3 \\
\hline
1,500 - 2,000 & 1 \\
\hline
2,000 - 3,000 & 2 \\
\hline
3,000 - 4,000 & 5 \\
\hline
4,000 - 5,000 & 4 \\
\hline
5,000 - 10,000 & 8 \\
\hline
10,000 - 50,000 & 24 \\
\hline
50,000 - 100,000 & 5 \\
\hline
100,000 - 500,000 & 13 \\
\hline
500,000 - 1,000,000 & 0 \\
\hline 
$\ge$ 1,000,000 & 3 \\
\hline 
\bottomrule
\end{tabular}
\caption{\label{Table: NLP_datasets_sizes} Dataset size at ACL and EMNLP in 2022.}
\end{table}
We have summarized the size of datasets in Table~\ref{Table: NLP_datasets_sizes} from which it is evident that most datasets have data samples in the range of 10,000 to 50,000. 

\section{Collaboration in Dataset Construction}
\label{sec: Collaboration}
It is important to list all the affiliations of authors at both conferences to understand the level of collaboration, especially between academic and industry researchers.
\begin{table*}[t!]
\centering
\scriptsize 
\begin{tabular}{  p{15.5cm}   }
\toprule
\hline 
{\bf  \centerline{Author Affiliation at ACL 2022}   }  \\ 
 \hline
University of Michigan, Singapore University of Technology and Design, Maastricht University, Carnegie Mellon University, Google Research, Harbin Institute of Technology, University of California Irvine, IBM Research, Tencent, Rensselaer Polytechnic Institute, University of Washington, University of Notre Dame, Syracuse University, Hong Kong Univ. of Sci and Tech, Georgia Institute of Technology, University of California Los Angeles, Columbia University, IBM Research Ireland, Université Paris-Saclay, ObTIC, Sorbonne Université, Université de Lorraine, Shanghai Jiao Tong University, Microsoft Research Asia, Xi’an Jiaotong University, DAMO Academy Alibaba Group, The Chinese University of Hong Kong, National University of Singapore, Tsinghua University, Cloud BU Huawei Technologies, ETH Zurich, Noah’s Ark Lab Huawei Technologies, Nanyang Technological University, University of Copenhagen,  Universitat Hamburg, Bucerius Law School, Stanford Law School, Athens University of Economics and Business, University of Sheffield, Chicago Kent College of Law, Beijing Jiaotong University, University of Tokyo, RIKEN AIP, Yahoo Japan, Toyota Technological Institute at Chicago, Duke University, University of Science and Technology of China, MIT, Microsoft Research, Allen Institute for AI, Microsoft, University of Waterloo, Fudan University, Shopee-Singapore, Vipshop (China) Co.Ltd, Fudan-Aishu Cognitive Intelligence Joint Research Center, Technical University of Darmstadt, IIT Delhi, Babylon Health, University of Aberdeen, University of Manchester, Soochow University, Chinese Academy of Social Sciences, University of Essex, Baidu Inc, Dartmouth College, California Institute of Technology, UC Santa Barbara, Vector Institute, Beijing Academy of Artificial Intelligence, Beijing Powerlaw Intelligent Technology Co, LMU Munich, Microsoft Azure AI, WeChat AI Tencent, Tencent AI Lab  \\
\hline 
\bottomrule
\end{tabular}
\caption{\label{Table: Author_affiliation_ACL}  At ACL 2022, there were 66 unique institutions to which dataset authors are affiliated.}
\end{table*}
\begin{table*}[t!]
\centering
\scriptsize
\begin{tabular}{ p{15.5cm}   }
\toprule
\hline 
{\bf  \centerline{Author Affiliation at EMNLP 2022}   }  \\ 
 \hline
Google Research, Bar Ilan University, Intel Labs Israel, Salesforce AI Research, Shanghai Mental Health Center, University of Cambridge, AWS AI Labs, Yonsei University, Seoul National University, University of Warwick, Northeastern University, AstraZeneca, King’s College London, The Alan Turing Institute,  Institut Polytechnique de Paris, Télécom-Paris, École Polytechnique, Korea University, Adobe Research, AIGEN Sciences, Hong Kong University of Science and Technology, University of Science and Technology of China,  Microsoft Azure Cognitive Services Research, Heidelberg University, SRH Hochschule Heidelberg, Beijing University of Posts and Telecommunications, Soochow University, Fudan University, Baidu Inc, University of Trento, Free University of Bozen-Bolzano, Fondazione Bruno Kessler, University of Dayton Research Institute, SIL International, Masakhane, Carnegie Mellon University, Microsoft Semantic Machines, University of Utah, Meta AI, University of Oregon, Adobe Research, UC San Diego, UMass Amherst, DAMO Academy Alibaba Group, Singapore University of Technology and Design, Hong Kong Polytechnic University, KAIST, SoftlyAI Research,   IIT Kharagpur, Goldman Sachs Data Science and ML Group, Leibniz University of Hannover, IIT Patna, Bosch Center for Artificial Intelligence, Heidelberg University, Shanghai AI Laboratory, Huawei Noah’s Ark Lab, University of Washington, The University of Hong Kong, JD AI Research, Umeå University, Adlede AB, Zhejiang University, AZFT Joint Lab for Knowledge Engine, UNC Chapel Hill, Allen Institute for AI, Tencent AI Lab, Renmin University of China, Hong Kong Polytechnic University, Chinese Academy of Sciences, University of Chinese Academy of Sciences, Islamic University of Technology, Chinese University of Hong Kong, Tsinghua University, IT University of Copenhagen, LMU Munich, Munich Center for ML, Shanghai AI Lab, University of Melbourne, RMIT University, MBZUAI, Data Friendly Space, ToggleCorp Solutions, Johannes Kepler University Linz, Stony Brook University, University of Arizona, Stanford University, VinAI Research, Hanoi University of Science and Technology, University of Oxford, University of Texas at Austin, UC Santa Barbara, Southeast University, Nanjing Origin Information Technology Company, IIT Bombay, Uniphore, HuggingFace, Yale University, University of Southern California, Kakao Corp, Soongsil University, Seoul National University, Virginia Tech, Jamia Millia, University of Sydney, University of New South Wales, University of Notre Dame, University of Groningen \\
\hline 
\bottomrule
\end{tabular}
\caption{\label{Table: Author_affiliation_EMNLP}  At EMNLP 2022, there were 103 unique institutions to which dataset authors are affiliated.}
\end{table*}
Table~\ref{Table: Author_affiliation_ACL} and Table~\ref{Table: Author_affiliation_EMNLP} indicate all the institutions, academic or industry, to which the authors are affiliated. 

The big winners in academia include Tsinghua University, the University of Washington, the Singapore University of Technology and Design, the National University of Singapore, Nanyang Technological University, the Hong Kong University of Science and Technology, and Stanford. 

The industry's big winners include Microsoft Research, Adobe Research, Google Research, Huawei Noah's Ark Lab, Alibaba DAMO Academy, and Tencent AI Lab.
\paragraph{Collaboration}There are 26 papers in which all the authors belong to \textit{one} institution, either academic or industry. Moreover, the number of documents in which all the authors are from \textit{one} academic institution is 20. 

Most papers consist of authors hailing from both academia and industry. This points to the advantages enjoyed when industry and academia researchers collaborate. It re-emphasizes long-held beliefs that the industrial labs provide the practical use cases, large-scale data, computing resources, and funds to foot the costs necessary to build a new dataset. In contrast, academia provides theoretical insights, novel methodologies, and expertise in meticulous experimental design. Out of the forty-three papers in which the industry and academic researchers collaborated, the number of academic \textit{last authors} is 20 while industry \textit{last authors} is 23. There are also few academia-academia collaborations. 

\section{Major Baselines}
\label{sec: Major_Baselines}
We discuss contemporary baselines for major tasks. Full details in Table~\ref{Table: NLP_Baselines}.
\begin{table}[!t]
\centering
\scriptsize  
\begin{tabular}{  p{2.8cm} | p{4cm} }
\toprule
\hline 
{\bf NLP Task}  &  {\bf Baselines   }  \\ 
\hline
 Question Answering & BERT, DistilBERT, BART \\
\hline
Abstractive Summarization & Transformer, BART, BM25, T5, PEGASUS, Longformer \\
\hline 
Extractive Summarization &  BERT, RoBERTa, Longformer \\
\hline 
Event Extraction & Pipeline approach: OneIE, FourIE; and for Joint approach: BERT, RoBERTa \\
\hline 
Relation Extraction & BERT, RoBERTa, SciBERT, PubmedBERT, BioBERT, BlueBERT \\
\hline 
Coreference Resolution, Event Coreference& RoBERTa, Deep Passage Retrieval \\
\hline
Text Generation, NLG & LexRank, TextRank, MMR, LSTM-CRF, BART \\
\hline 
Hate Speech Detection  & HateBert, ToxDectRoBERTa and LSTM, BERT, RoBERTa, DeBERTa \\
\hline
Natural Language Understanding & TFIDF-SVM, BERT, RoBERTa, DeBERTa, Longformer, BigBird \\
\hline
Named Entity Recognition & BERT, RoBERTa\\
\hline
Machine translation & mBERT, mBART, mT5, OPUS-MT \\
\hline
Image captioning & ViT, mT5, CIDEr, InceptionV3 \\
\hline
\bottomrule
\end{tabular}
\caption{\label{Table: NLP_Baselines} The major baselines from papers in our study.}
\end{table}
\section{The Rise of Multimodality}
\label{sec: Rise_of_Multimodality}
The growing demand for NLP systems that accept at least two modalities as input, for example, \textit{text} and \textit{image}, leads to a further increase in demand for multimodal datasets.

Such multimodal datasets make it possible to meet the needs for both textual and visual embeddings when training \textit{visual-language} systems for tasks such as multimodal dialogue summarization, multimodal response generation, visual question answering, multimodal dialogue sentiment analysis, visual storytelling, visual entity linking, multimodal information extraction, among other things.

Papers which introduced multimodal datasets include: MSCTD~\citep{liang-etal-2022-msctd}, WikiDiverse~\citep{wang-etal-2022-wikidiverse}, DuReader\_vis~\citep{qi-etal-2022-dureadervis}, XFUND~\citep{xu-etal-2022-xfund}, Crossmodal-3600~\citep{thapliyal-etal-2022-crossmodal}, FigMemes~\citep{liu-etal-2022-figmemes},~\citep{leong-etal-2022-bloom}, JDDC 2.1~\citep{zhao-etal-2022-jddc}, ~\citet{sun-etal-2022-visual}, and~\citep{thapa-etal-2022-multi}.

MSCTD~\citep{liang-etal-2022-msctd} consists of 142,871 English-Chinese utterance pairs and supports the development of multimodal chat translation and dialogue sentiment analysis. WikiDiverse~\citep{wang-etal-2022-wikidiverse} contains 8K image captions and makes evaluating multimodal entity linking (MEL) systems possible. DuReader\_vis~\citep{qi-etal-2022-dureadervis} includes 15K QA pairs in Chinese and 158K document images, making evaluating visual question answering (VQA) systems possible. XFUND~\citep{xu-etal-2022-xfund} introduced 1,393 forms and supports multilingual document understanding, semantic entity recognition, and relation extraction for seven languages, that is, Chinese, Japanese, Spanish, French, Italian, German, and Portuguese. Crossmodal-3600~\citep{thapliyal-etal-2022-crossmodal} comprises 3,600 images and 261,373 image captions necessary to support the development of multilingual image captioning systems. FigMemes~\citep{liu-etal-2022-figmemes}, which contains 5,141 labels, makes it possible to identify figurative language in politically-opinionated memes. BLOOM~\citep{leong-etal-2022-bloom} is a massive dataset that covers 363 languages and supports the development of image captioning, visual storytelling, speech synthesis and recognition, and language modeling. On the other hand, JDDC 2.1~\citep{zhao-etal-2022-jddc} enables the development of query writing, response generation, discourse parsing, and summarization tasks in Chinese. MovieUN~\citep{zhang-etal-2022-movieun} enables movie understanding in Chinese. To allow the development of visual entity linking systems, \newcite{sun-etal-2022-visual} developed a new dataset. \newcite{thapa-etal-2022-multi}'s dataset enables hate speech detection in natural language.

\section{Quantifying the Language Gap: Monolingual vs. Multilingual Data}
\label{sec: Language_Gap_Multilingual}
Unsurprisingly, most datasets published at ACL and EMNLP conferences contain English text. However, there is a notable number of non-English datasets, too. To study the multilingual aspect of these datasets, we categorized the datasets into (i) \emph{Non-English and Monolingual}, which includes only datasets curated with data in one language and that language is not English; and (ii) \emph{Multilingual} in which the data is curated in more than one language. 

We have provided further details for all the datasets in the Appendix~\ref{sec: Dataset_details}.
\begin{figure}[t!]
\centering
\begin{tikzpicture}
\begin{axis}[
  ybar,
  ylabel={ {\bf Count.} },
  xlabel={ {\bf Language in the dataset.} },
  symbolic x coords={Bangla, French, Korean, Patois*, Chinese},
  xtick=data,
  ymin=0, ymax=10,
  ytick={0,10},
  legend pos=north west,
  ymajorgrids=true,
  grid style=dashed,
  width=7.5cm,
  height=3cm,
]
\addplot[
  color=blue,
  fill=red!30!white,
  nodes near coords,
  ]
  coordinates {
    (Bangla, 1)
    (French, 2)
    (Korean, 2)
    (Patois*, 1)
    (Chinese, 8)
  };
\legend{Monolingual data}
\end{axis}
\end{tikzpicture}
\caption{The number of non-English monolingual datasets. \textit{Patois*} refers to \textit{Jamaican Patois}.}
\label{fig: Non_English_and_Monolingual}
\end{figure}
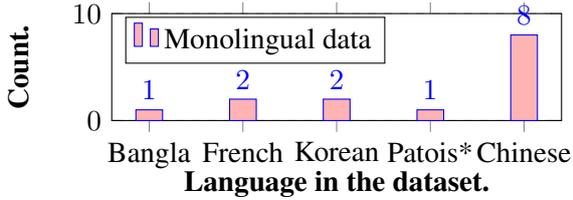
\paragraph{Non-English and Monolingual} This category includes datasets containing only one language and that language is not English. Refer to Table~\ref{Table: Non_English_and_Monolingual}.

The five languages represented in this category are: \textbf{French}—French CrowS-Pairs~\citep{neveol-etal-2022-french} and~\newcite{louis-spanakis-2022-statutory}; \textbf{Bangla}—BanglaRQA~\citep{ekram-etal-2022-banglarqa}; \textbf{Korean}—KOLD~\citep{jeong-etal-2022-kold} and APEACH~\citep{yang-etal-2022-apeach}; \textbf{Jamaican Patois}—JamPatoisNLI~\citep{armstrong-etal-2022-jampatoisnli}; \textbf{Chinese}—~\newcite{jiang-etal-2022-chinese}, DuReader\_vis~\citep{qi-etal-2022-dureadervis}, LEVEN~\citep{yao-etal-2022-leven}, D4~\citep{yao-etal-2022-d4}, Title2Event~\citep{deng-etal-2022-title2event}, DuQM~\citep{zhu2022duqm}, ~\citep{zhao-etal-2022-fine-grained}, JDDC 2.1~\citep{zhao-etal-2022-jddc}, MovieUN~\citep{zhang-etal-2022-movieun} and ~\citep{zhou-etal-2022-towards-identifying}. 

Moreover,~\newcite{maheshwari-etal-2022-benchmark} introduced a benchmark dataset for \textbf{Sanskrit}. 
\begin{table}[!t]
\centering
\footnotesize 
\begin{tabular}{  p{4cm} | p{2cm} }
\toprule
\hline 
{\bf Monolingual Dataset}  &  {\bf Language }  \\ 
\hline
 French CrowS-Pairs~\citep{neveol-etal-2022-french} and ~\newcite{louis-spanakis-2022-statutory}  & French \\
\hline
 BanglaRQA~\citep{ekram-etal-2022-banglarqa} & Bangla \\
 \hline 
 ~\newcite{maheshwari-etal-2022-benchmark} & Sanskrit \\
 \hline
 KOLD~\citep{jeong-etal-2022-kold} and APEACH~\citep{yang-etal-2022-apeach}   & Korean \\
 \hline
 JamPatoisNLI~\citep{armstrong-etal-2022-jampatoisnli} & Jamaican Patois \\
 \hline
 \newcite{jiang-etal-2022-chinese}, DuReader\_vis~\citep{qi-etal-2022-dureadervis}, D4~\citep{yao-etal-2022-d4}, Title2Event~\citep{deng-etal-2022-title2event},  DuQM~\citet{zhu2022duqm};~\citep{zhao-etal-2022-fine-grained}, JDDC 2.1~\citep{zhao-etal-2022-jddc}, MovieUN~\citep{zhang-etal-2022-movieun}, ~\citep{zhou-etal-2022-towards-identifying}  & Chinese \\
\hline 
\bottomrule
\end{tabular}
\caption{\label{Table: Non_English_and_Monolingual} The Non-English but \textit{monolingual} datasets.   }
\end{table}
\begin{table*}[!t]
\tiny
\centering
\resizebox{\textwidth}{!}{\begin{tabular}{c|c|c|c|c|c|c|c|c|c|c|c}
\toprule
\hline 
\textbf{Data Source} & Tests & Legal & Govt & Book & Crowdsourcing & Online & Movie,TV & Prompting & News & Med. & OldData\\
\hline
\textbf{Count}   & 1  & 3 & 2 & 4 & 15 & 31 & 2 & 2 & 7 & 3 & 6\\
\hline 
\bottomrule
\end{tabular}}
\caption{\label{Table:DataSources}Sources of text, image, and video used to construct datasets. Test* refers to language test materials. Legal* refers to legal documents e.g., law articles and patents. Govt means government websites. Online* includes forums, GitHub, e-commerce platforms, search engines, and social media. Med* refers to medical reports and biomedical literature. oldData* refers to existing datasets modified to create new datasets.}
\end{table*}
\paragraph{Multilingual} These datasets contain multiple languages (see Table~\ref{Table: Multilingual}). MSCTD~\citep{liang-etal-2022-msctd} in English and Chinese;~\newcite{filighera-etal-2022-answer} in English and German; DiS-ReX~\citep{bhartiya-etal-2022-dis} in English, French, Spanish and German; DivEMT \citep{sarti-etal-2022-divemt} in Arabic, Dutch, Italian, Turkish, Ukrainian, and Vietnamese; ~\newcite{maronikolakis-etal-2022-listening} in Brazilian Portuguese, German, Hindi, Swahili and English; XFUND~\citep{xu-etal-2022-xfund} in 7 languages Chinese, Japanese, Spanish, French, Italian, German and Portuguese; Crossmodal-3600~\citep{thapliyal-etal-2022-crossmodal} in 36 languages; MT-GenEval~\citet{currey-etal-2022-mt} in Arabic, French, German, Hindi, Italian, Portuguese, Russian and Spanish; IndicNLG Benchmark~\citet{kumar-etal-2022-indicnlg} in 11 Indic languages; EUR-Lex-Sum~\citep{aumiller-etal-2022-eur} in 24 official European languages; ClidSum~\citep{wang-etal-2022-clidsum} in English, German and Chinese; BLOOM~\citep{leong-etal-2022-bloom} in 363 languages; MEE~\citep{pouran-ben-veyseh-etal-2022-mee} in 8 languages English, Spanish, Portuguese, Polish, Turkish, Hindi, Korean, and Japanese; HumSet~\citep{fekih-etal-2022-humset} in English, Spanish, French; ~\citet{lai-etal-2022-multilingual-subevent}  in 5 languages English, Danish, Spanish, Turkish, and Urdu; TyDiP~\citep{srinivasan-choi-2022-tydip} in 9 languages Hindi, Korean, Spanish, Tamil, French, Vietnamese, Russian, Afrikaans, Hungarian; and ClassBases~\citep{zavarella-etal-2022-tracking, hurriyetoglu-etal-2021-multilingual} in English, Spanish, Portuguese. 
\begin{table}[!t]
\centering
\scriptsize 
\begin{tabular}{  p{2cm} | p{0.8cm} | p{3.5cm} }
\toprule
\hline 
{\bf Multilingual Dataset}  & {\bf \# Lang.} &  {\bf Language }  \\ 
\hline
MSCTD~\citep{liang-etal-2022-msctd} & 2 & English, Chinese \\
\hline
\citet{filighera-etal-2022-answer} & 2 & English, German \\
\hline
ClidSum~\citep{wang-etal-2022-clidsum} & 3 & English, German, Chinese \\
\hline
ClassBases~\citep{zavarella-etal-2022-tracking, hurriyetoglu-etal-2021-multilingual} & 3 & English, Spanish, Portuguese \\
\hline
HumSet~\citep{fekih-etal-2022-humset} & 3 &  English, Spanish, French \\
\hline 
DiS-ReX~\citep{bhartiya-etal-2022-dis} & 4 & English, French, Spanish, German \\
\hline 
\citet{maronikolakis-etal-2022-listening} & 5 & Brazilian Portuguese, German, Hindi, Swahili, English \\
\hline
\citet{lai-etal-2022-multilingual-subevent} & 5  &  English, Danish, Spanish, Turkish, Urdu \\
\hline
DivEMT \cite{sarti-etal-2022-divemt} & 6 & Arabic, Dutch, Italian, Turkish, Ukrainian, Vietnamese \\
\hline 
XFUND~\citep{xu-etal-2022-xfund} & 7 & Chinese, Japanese, Spanish, French, Italian, German, Portuguese \\
\hline
MEE~\citep{pouran-ben-veyseh-etal-2022-mee} & 8 & English, Spanish, Portuguese, Polish, Turkish, Hindi, Korean, Japanese \\
\hline
MT-GenEval~\citep{currey-etal-2022-mt} & 8 & Arabic, French, German, Hindi, Italian, Portuguese, Russian, Spanish \\
\hline
TyDiP~\citep{srinivasan-choi-2022-tydip} &  9  &  Hindi, Korean, Spanish, Tamil, French, Vietnamese, Russian, Afrikaans, Hungarian \\
\hline
IndicNLG Benchmark~\citep{kumar-etal-2022-indicnlg} & 11  & Assamese, Bengali, Gujarati, Hindi, Marathi, Odiya, Punjabi, Kannada, Malayalam, Tamil, Telugu \\
\hline
EUR-Lex-Sum~\citep{aumiller-etal-2022-eur} & 24  & Bulgarian, Croatian, Czech, Danish, Dutch, English, Estonian, Finnish, French, German, Greek, Hungarian, Irish, Italian, Latvian, Lithuanian, Maltese, Polish, Portuguese, Romanian, Slovak, Slovenian, Spanish, Swedish\\
\hline
Crossmodal-3600~\citep{thapliyal-etal-2022-crossmodal} & 36 & Arabic, Bengali, Chinese-Simplified, Croatian, Cusco Quechua, Czech, Danish, Dutch, English, Filipino, Finnish, French, German, Greek, Hebrew, Hindi, Hungarian, Indonesian, Italian, Japanese, Korean, Maori, Norwegian, Persian, Polish, Portuguese, Romanian, Russian, Spanish, Swahili, Swedish, Telugu, Thai, Turkish, Ukrainian, Vietnamese. \\
\hline
BLOOM~\citet{leong-etal-2022-bloom} & 363 & \textit{Full list of 363 languages in Appendix \ref{sec: BLOOM_Dataset}.} \\
\hline 
\bottomrule
\end{tabular}   
\caption{\label{Table: Multilingual} Multilingual datasets in order of increasing number of languages. }
\end{table}

\section{Source of Data, Dataset Creation}
\label{sec: SourceOfData}
A thorough analysis of the sources of texts and images (see Table \ref{Table:DataSources}) used to construct the datasets reveals the following. 1) Diverse \textbf{online resources} play a critical role in availing the texts, images, and videos leveraged to construct datasets. The online sources vary from GitHub repositories, online forums, search engines and e-commerce, e.g., Baidu and JD.com; social media platforms, e.g., Reddit and Twitter; digital book collections; video and TV streaming platforms, e.g., YouTube and Xigua Video, and NGO and government websites. 2) \textbf{Wikipedia and news articles} remain a profound source of texts needed to construct new corpora. 3) The \textbf{materials} used to construct datasets include English language exams, patents, news articles, textbooks, medical reports, image collections, and movie clips. 4) \textbf{Crowdsourcing}, which involves recruiting human annotators, remains a relevant method to collect data. 5) While most datasets introduce entirely new data, some authors re-purpose \textbf{existing datasets}, creating new corpora that suit new tasks or analyses. 6) We note that \textbf{prompting large language models (LLMs)} to generate training examples has emerged as a new method to overcome the time and budget constraints that come with recruiting human annotators. 
\section{Conclusion}
\label{sec: Conclusion}
Our study focuses exclusively on datasets and NLP methodologies presented at the ACL and EMNLP conferences in 2022, selected for their outstanding quality and influential insights in natural language processing. These conferences are recognized for setting trends and directions in NLP research, ensuring that our analysis reflects the forefront of the discipline. Despite the narrow scope, we anticipate our findings will offer substantial value by highlighting innovative datasets introduced at these venues, serving as a vital resource and guide for researchers aiming to develop new datasets in NLP.
\section*{Acknowledgements}
We thank Iqra Ali from the NAIST-NLP group for helping with gathering information about the data sources.
\newpage
\bibliography{main}
\bibliographystyle{acl_natbib}

\newpage
\appendix
\label{appendix}
\section{Dataset Size}
\begin{figure*}[!t]
\centering
\begin{tikzpicture}
\begin{axis}[
  ybar,
  ylabel={Count},
  xlabel={Dataset Size},
  symbolic x coords={0.5K, 1K, 1.5K, 2K, 3K, 4K, 5K, 10K, 50K, 100K, 500K, 1M, >1M },
  xtick=data,
  ymin=0, ymax=32,
  ytick={0,8,16,24,32},
  legend pos=north west,
  ymajorgrids=true,
  grid style=dashed,
  width=16cm,
  height=3cm
]
\addplot[
  color=blue,
  fill=blue!30!white,
  nodes near coords,
  ]
  coordinates {
    (0.5K, 3)
    (1K, 2)
    (1.5K, 3)
    (2K, 1)
    (3K, 2)
    (4K, 5)
    (5K, 4)
    (10K, 8)
    (50K, 24)
    (100K, 5)
    (500K, 13)
    (1M, 0)
    (>1M, 3)
  };
\legend{Dataset size}
\end{axis}
\end{tikzpicture}
\caption{The varying sizes of datasets. Most datasets tend to have between 10K—50K samples.}
\label{fig: Dataset_size}
\end{figure*}
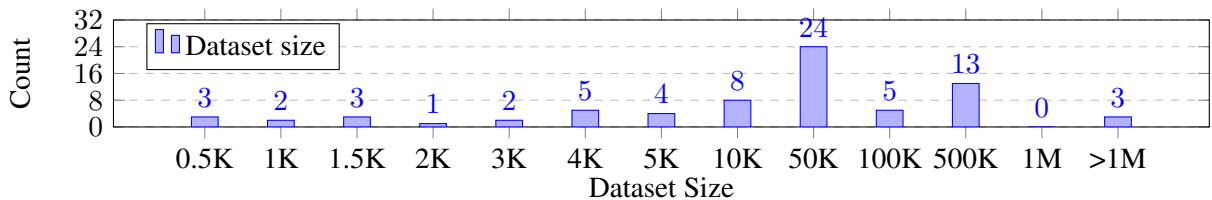
\section{Dataset details}
\label{sec: Dataset_details}
Details for all the datasets surveyed. Due to a space limitation, we split the table into seven parts. Therefore, the Tables \ref{Table: Datasets_summary_Part_I}, \ref{Table: Datasets_summary_Part_II}, \ref{Table: Datasets_summary_Part_III}, \ref{Table: Datasets_summary_Part_IV}, \ref{Table: Datasets_summary_Part_V}, \ref{Table: Datasets_summary_Part_VI}, \ref{Table: Datasets_summary_Part_VII} are a presented in a continuous manner. 
\begin{table*}[!t]
\centering
\scriptsize 
\begin{tabular}{ p{4cm}| p{2cm} | p{2cm} | p{3cm} | p{1cm} | p{1cm}   }
\toprule
\hline 
 Paper/ Dataset & Size (\#sentences) & NLP Topic & Baselines & Multi-lingual? & Multi-modal? \\
\midrule
{\bf Track: ACL 2022} \\
 \hline
~\citet{ghosal-etal-2022-cicero} CICERO: A Dataset for Contextualized Commonsense Inference in Dialogues& 53,105 inst. & dialogue reasoning & GLUCOSE-T5, RoBERTa-large, ELECTRA-large,
T5-large, Unified QA Large & No & No \\  
\hline
~\citet{louis-spanakis-2022-statutory} A Statutory Article Retrieval Dataset in French & 1,100 & Statutory article retrieval & TF-IDF, BM25, word2vec, fastText, RoBERTa, CamemBERT & No (French) & No \\ 
\hline
~\citet{sun-etal-2022-conditionalqa} ConditionalQA: A Complex Reading Comprehension Dataset with Conditional Answers & 3,427 questions & Reading Comprehension, QA & ETC, DocHopper, FiD, Human & No & No \\
\hline
~\citet{du-etal-2022-e} e-CARE: a New Dataset for Exploring Explainable Causal Reasoning &21,000 &Causal Reasoning &BERT, RoBERTa, XLNet, ALBERT, GPT2, BART &No &No \\
\hline
~\citet{xu-etal-2022-fantastic} Fantastic Questions and Where to Find Them: FairytaleQA -- An Authentic Dataset for Narrative Comprehension &10,580 questions &Question answering &BERT, DistilBERT, BART &No &No \\
\hline
~\citet{neveol-etal-2022-french} French CrowS-Pairs: Extending a challenge dataset for measuring social
bias in masked language models to a language other than English &1,679 sentence pairs &Bias &CamemBERTbase, FlauBERTbase, FrALBERT &No (French) &No \\
\hline
~\citet{cheng-etal-2022-hitab} HiTab: A Hierarchical Table Dataset for Question Answering and Natural Language Generation &3,597 tables &QA, NLG &Neural Symbolic Machine, TaPas, MML, REINFORCE, MAPO &No &No \\
\hline
~\citet{cheng-etal-2022-iam} IAM: A Comprehensive and Large-Scale Dataset for Integrated Argument Mining Tasks &69,666 sentences &Argument Mining &RoBERTa-base &No &No \\
\hline
~\citet{cao-etal-2022-kqa} KQA Pro: A Dataset with Explicit Compositional Programs for Complex Question Answering over Knowledge Base &117,970  sentences &QA &KVMemNet, SRN, EmbedKGQA, RGCN, BART &No &No \\
\hline
~\citet{chalkidis-etal-2022-lexglue} LexGLUE: A Benchmark Dataset for Legal Language Understanding in English &Combination of existing datasets &Legal NLU &TFIDF-SVM, BERT, RoBERTa, DeBERTa, Longformer, BigBird, Legal-BERT, CaseLaw-BERT &No & No \\
\hline 
~\citet{liang-etal-2022-msctd} MSCTD: A Multimodal Sentiment Chat Translation Dataset &142,871 English-Chinese utterance pairs &Multimodal chat translation; dialogue sentiment analysis & many &Yes (English, Chinese) &Yes\\
\hline
~\citet{kamezawa-etal-2022-rnsum} RNSum: A Large-Scale Dataset for Automatic Release Note Generation via Commit Logs Summarization &82,000 English release notes &Release Note Generation &Glyph, Clustering, BERT, CodeBERT, BART &No &No \\
\hline
~\citet{chen-etal-2022-summscreen} SummScreen: A Dataset for Abstractive Screenplay Summarization &26,900 instances &Abstractive Summarization &Longformer, standard Transformer, BART-large, BM25, ROUGE &No &No \\
\hline
~\citet{wang-etal-2022-textomics} Textomics: A Dataset for Genomics Data Summary Generation &22,273 pairs of genomics data &Genomics Data Summary Generation &kNN-Vec2Text, SPECTER, SciBERT, BioBERT, SentBERT, BERT, SentBERT &No &No \\
\hline 
~\citet{hartvigsen-etal-2022-toxigen} TOXIGEN: A Large-Scale Machine-Generated Dataset for Adversarial and Implicit Hate Speech Detection &274k toxic and benign statements &Hate Speech Detection &HateBert, ToxDectRoBERTa &No &No \\
\hline
~\citet{khan-etal-2022-watclaimcheck} WatClaimCheck: A new Dataset for Claim Entailment and Inference &33,721 claims &Automated fact checking &TF-IDF, BERT,  Bi-LSTM and Bi-GRU, hierarchical attention networks, RoBERTa-base &No &No \\
\hline
 ~\citet{wang-etal-2022-wikidiverse} WikiDiverse: A Multimodal Entity Linking Dataset with Diversified Contextual Topics and Entity Types &8K captions &Multimodal Entity Linking, NER &BM25, WikiVec, BLINK, CLIP, BERT, UNITER, UNITER*, LXMERT &No &Yes \\
 \hline 
\bottomrule
\end{tabular}
\caption{\label{Table: Datasets_summary_Part_I} This table shows details for the name of the paper, the size of the dataset, the NLP topic of interest, baselines and multilingualism or multimodality status of the dataset (\texttt{Part I}). }
\end{table*}
\begin{table*}[!t]
\centering
\scriptsize
\begin{tabular}{ p{4cm}| p{2cm} | p{2cm} | p{3cm} | p{1cm} | p{1cm}   }
\toprule
\hline 
 Paper/ Dataset & Size (\#sentences) & NLP Topic & Baselines & Multi-lingual? & Multi-modal? \\
\midrule
{\bf Track: ACL 2022 continued} \\
\hline
 ~\citet{wang-etal-2022-wikidiverse} WikiDiverse: A Multimodal Entity Linking Dataset with Diversified Contextual Topics and Entity Types &8K captions &Multimodal Entity Linking, NER &BM25, WikiVec, BLINK, CLIP, BERT, UNITER, UNITER*, LXMERT &No &Yes \\
 \hline
 ~\citet{filighera-etal-2022-answer} Your Answer is Incorrect... Would you like to know why? Introducing a Bilingual Short Answer Feedback Dataset &4,519 submissions &Explainable feedback systems &T5-base, mT5-base models &Yes (English, German) &No \\
\hline 
~\citet{bhartiya-etal-2022-dis} DiS-ReX: A Multilingual Dataset for Distantly Supervised Relation Extraction &1.8 million sentences &Multilingual relation extraction &mBERT &Yes (English, French, Spanish, German) & No\\
\hline
~\citet{papadopoulos-korfiatis-etal-2022-primock57} PriMock57: A Dataset Of Primary Care Mock Consultations &56 mock consultations &Automatic Speech Recognition &ASR Benchmarks (Kaldi, NeMo QuartzNet, Conformer, Google Cloud Speech-to-text, Amazon Transcribe Medical, Azure Speech-to-text) Consultation Note Generation Benchmarks (BART-CNN, BERT-ext, Random, BART-finet) &No &No \\
\hline
~\citet{tavella-etal-2022-wlasl} WLASL-LEX: a Dataset for Recognising Phonological Properties in American Sign Language &Sign language recognition & -- & MLP, LSTM, GRU,  I3D 3D Convolutional Neural Network, Spatio-Temporal Graph Convolutional Network &No &No \\
\midrule
{\bf Track: Findings of ACL 2022} \\
\hline
~\citet{jiang-etal-2022-chinese} Chinese Synesthesia Detection: New Dataset and Models &7,217 sentences &synesthesia detection &BiLSTM+CRF, BERT+CRF, SR-BiLSTM, PF-BERT &No (Chinese) &No \\
\hline
~\citet{qi-etal-2022-dureadervis} DuReader\_vis: A Chinese Dataset for Open-domain Document Visual Question Answering &15K QA pairs; 158,000 document images &Visual Question Answering &DocVRE->BM25, DocVQA->RobertaXLM-base, BERT-base-Chinese &No (Chinese) &Yes \\
\hline
~\citet{ma-etal-2022-encbp} EnCBP: A New Benchmark Dataset for Finer-Grained Cultural Background Prediction in English
18K documents &cultural background prediction & many &No &No \\
\hline
~\citet{nakamura-etal-2022-hybridialogue} HybriDialogue: An Information-Seeking Dialogue Dataset Grounded on Tabular and Textual Data &4,844 dialogues & Dialogue & Okapi BM25 retriever, SentenceBERT, TaPas, DialoGPT & No & No \\
\hline
~\citet{yao-etal-2022-leven} LEVEN: A Large-Scale Chinese Legal Event Detection Dataset & 63,616 sentences & Legal Event Detection & BiLSTM, BiLSTM+CRF,  BERT, BERT+CRF, DMCNN, DMBERT & No (Chinese) & No \\
\hline
~\citet{maronikolakis-etal-2022-listening} Listening to Affected Communities to Define Extreme Speech: Dataset and Experiments & 20,297 social media passages & hate speech detection & mBERT, XLM-R, langBERT & Yes (Brazilian Portuguese, German, Hindi, Swahili, English) & No \\
\hline
\bottomrule
\end{tabular}
\caption{\label{Table: Datasets_summary_Part_II} This table shows details for the name of the paper, the size of the dataset, the NLP topic of interest, baselines and multilingualism or multimodality status of the dataset (\texttt{Part II}). }
\end{table*}
\begin{table*}[!t]
\centering
\scriptsize
\begin{tabular}{ p{4cm}| p{2cm} | p{2cm} | p{3cm} | p{2cm} | p{1cm}   }
\toprule
\hline 
 Paper/ Dataset & Size (\#sentences) & NLP Topic & Baselines & Multi-lingual? & Multi-modal? \\
\midrule
{\bf Track: Findings of ACL 2022 continued} \\
\hline
~\citet{shen-etal-2022-mred} MReD: A Meta-Review Dataset for Structure-Controllable Text Generation & 45k meta-review sentences & Controllable Text Generation & Extractive Baselines (LexRank, TextRank, MMR, n LSTM-CRF, bart-large-cnn) & No & No \\
\hline
~\citet{sun-etal-2022-safety} On the Safety of Conversational Models: Taxonomy, Dataset, and Benchmark & 11,492 utterances & Dialogue safety & DialoGPT, Blenderbot, Plato-2, PerspectiveAPI, Detoxify, Build it, Break it Fix it, BotAdversarial Dialogue Safety Classifier  & No & No \\
\hline
~\citet{xu-etal-2022-xfund} XFUND: A Benchmark Dataset for Multilingual Visually Rich Form Understanding & 1,393 forms & multilingual document understanding, Semantic Entity Recognition, Relation Extraction & LayoutXLM, XLM-RoBERTa, InfoXLM & 7 languages (Chinese, Japanese, Spanish, French, Italian, German, Portuguese) & Yes \\
\midrule
{\bf Track: EMNLP 2022 } \\
\hline
~\citet{thapliyal-etal-2022-crossmodal} Crossmodal-3600: A Massively Multilingual Multimodal Evaluation Dataset & 3600 images; 261,375 captions & multilingual image captioning
& ViT, mT5, CIDEr & 36 languages & Yes \\
\hline
~\citet{eirew-etal-2022-cross} Cross-document Event Coreference Search: Task, Dataset and Modeling & 2,772,968 passages & Coreference Resolution; event coreference & Deep Passage Retrieval (DPR) & No & No \\
\hline
~\citet{wang-etal-2022-maven} MAVEN-ERE: A Unified Large-scale Dataset for Event Coreference, Temporal, Causal, and Subevent Relation Extraction & 103,193 event coreference chains, 1,216,217 temporal relations, 57,992 causal relations, and 15,841 subevent relations & Event Coreference, Temporal, Causal, and Subevent Relation Extraction & RoBERTa-BASE & No & No \\
\hline
~\citet{reid-etal-2022-m2d2} M2D2: A Massively Multi-Domain Language Modeling Dataset & 8.5B tokens & Multi-Domain Language Modeling & 112M GPT2 & No & No \\
\hline
~\citet{wang-etal-2022-squality} SQuALITY: Building a Long-Document Summarization Dataset the Hard Way & 100 stories, 625 examples & Summarization & BART-large, Dense Passage Retriever, PEGASUS, Longformer Encoder-Decoder (LED-base), Human Evaluation & No & No \\
\hline
~\citet{laban-etal-2022-near} Near-Negative Distinction: Giving a Second Life to Human Evaluation Datasets & N/A & NLG evaluation & GLUE, GLGE, GEM, BARTScore,  & No & No \\
\hline
~\citet{yao-etal-2022-d4} D4: a Chinese Dialogue Dataset for Depression-Diagnosis-Oriented Chat & 1339 dialogues & Dialogue (Depression-Diagnosis-Oriented Chat) & BART, CPT, BERT & No (Chinese) & No \\
\hline
~\cite{liu-etal-2022-measuring} Measuring Context-Word Biases in Lexical Semantic Datasets & N/A & Model bias & BERT & No & No \\
\hline
~\citet{currey-etal-2022-mt} MT-GenEval: A Counterfactual and Contextual Dataset for Evaluating Gender Accuracy in Machine Translation & 300 sentences & Gender accuracy in translation & Gender Evaluation Benchmarks— WinoMT, MuST-SHE, GeBioCorpus, SimpleGEN. Model— Transformer-base & Yes (Arabic, French, German, Hindi, Italian, Portuguese, Russian, Spanish.) & No \\
\hline 
\bottomrule
\end{tabular}
\caption{\label{Table: Datasets_summary_Part_III} This table shows details for the name of the paper, the size of the dataset, the NLP topic of interest, baselines and multilingualism or multimodality status of the dataset (\texttt{Part III}). }
\end{table*}
\begin{table*}[!t]
\centering
\scriptsize
\begin{tabular}{ p{4cm}| p{2cm} | p{2cm} | p{3cm} | p{2cm} | p{1cm}   }
\toprule
\hline 
 Paper/ Dataset & Size (\#sentences) & NLP Topic & Baselines & Multi-lingual? & Multi-modal? \\
\midrule
{\bf Track: EMNLP 2022 continued} \\
\hline
~\citet{kim-etal-2022-botstalk} BotsTalk: Machine-sourced Framework for Automatic Curation of Large-scale Multi-skill Dialogue Datasets & 300K conversations & Dialogues, open-domain chatbots & 256-M parameter poly-encoder, BART,  Human Evaluation & No & No \\ 
\hline
~\citet{kumar-etal-2022-indicnlg} IndicNLG Benchmark: Multilingual Datasets for Diverse NLG Tasks in Indic Languages & 8M examples across 5 tasks (biography generation using Wikipedia infoboxes, news headline generation, sentence summarization, paraphrase generation and, question generation) and 11 languages & Multilingual NLG & IndicBART, mBART, YANMTT toolkit & Yes. 11 languages from o two language families i.e., Indo-Aryan and Dravidian {as, bn, gu,hi, kn, ml, mr, or, pa, ta, te} & No \\ 
\hline
~\citet{sun-etal-2022-phee} PHEE: A Dataset for Pharmacovigilance Event Extraction from Text & 5000 events & drug safety  & ACE for NER, EEQA for Extractive QA, BioBERT, SciFive for Generative QA, T5 & No & No \\
\hline
~\citet{guo-etal-2022-questioning} Questioning the Validity of Summarization Datasets and Improving Their Factual Consistency & 306,522 samples  & abstractive summarization & BART-base & No & No \\
\hline
~\citet{kim-etal-2022-simple} Simple Questions Generate Named Entity Recognition Datasets  & N/A & Framework to generate NER datasets & RoBERTa, BOND, TALLOR, GeNER & No & No \\
\hline
~\citet{deng-etal-2022-title2event} Title2Event: Benchmarking Open Event Extraction with a Large-scale Chinese Title Dataset  &  42,000 news titles in 34 topics &  Event extraction &  BERT &  No (Chinese) &  No \\
\hline
~\citet{liu-etal-2022-figmemes} FigMemes: A Dataset for Figurative Language Identification in Politically-Opinionated Memes & 5,141 labels & figurative language classification in politically-opinionated memes & SentenceBERT, Text only (BERT, DeBERTa) Image only (ConvNeXt, CLIP-CNN), Multimodal (CLIP, VinVL, BERT+CLIP, CLIP-MM-OOD) & No & Yes \\
\hline
~\citet{wang-etal-2022-paratag} ParaTag: A Dataset of Paraphrase Tagging for Fine-Grained Labels, NLG Evaluation, and Data Augmentation & 30k sentence pairs & Paraphrase identification and generation & (1) Paraphrasing detection — BERT, RoBERTa, DeBERTa-v3 (2) Paraphrasing generation — BART & No & No \\
\hline
~\citet{aumiller-etal-2022-eur} EUR-Lex-Sum: A Multi- and Cross-lingual Dataset for Long-form & Summarization in the Legal Domain & 1,500 document/summary pairs per language Extractive Summarization & Zero-shot Extractive Baselines—LexRank AND Cross-lingual Baselines—Longformer Encoder Decoder (LED) AND translation—OPUS-MT & Yes (24 official European languages) & No \\
\hline
~\citet{wang-etal-2022-clidsum} ClidSum: A Benchmark Dataset for Cross-Lingual Dialogue Summarization & 67k+ dialogues AND 112k+ summaries & Dialogue Summarization & mBART-50 & Yes (English, German and Chinese) & No \\
\hline
~\citet{zhu2022duqm} DuQM: A Chinese Dataset of Linguistically Perturbed Natural Questions for Evaluating the Robustness of Question Matching Models & 10,121 examples & Chinese Question Matching  & BERT-b, ERNIE-b, RoBERTa-b, MacBERT-b & No (Chinese) & No \\
\hline
~\citet{bonaldi-etal-2022-human} Human-Machine Collaboration Approaches to Build a Dialogue Dataset for Hate Speech Countering & 3059 dialogues & Hate Speech in Dialogue & DialoGPT, T5-2m, T5-1m & English & No \\
\hline
\bottomrule
\end{tabular}
\caption{\label{Table: Datasets_summary_Part_IV} This table shows details for the name of the paper, the size of the dataset, the NLP topic of interest, baselines and multilingualism or multimodality status of the dataset (\texttt{Part IV}). }
\end{table*}
\begin{table*}[!t]
\centering
\scriptsize
\begin{tabular}{ p{4cm}| p{2cm} | p{2cm} | p{3cm} | p{2cm} | p{1cm}   }
\toprule
\hline 
 Paper/ Dataset & Size (\#sentences) & NLP Topic & Baselines & Multi-lingual? & Multi-modal? \\
\midrule
{\bf Track: EMNLP 2022 continued} \\
\hline
~\citet{leong-etal-2022-bloom} Bloom Library: Multimodal Datasets in 300+ Languages for a Variety of Downstream Tasks & -- & Language modeling, Image captioning, Visual storytelling, and Speech synthesis/recognition & Language Modeling—DistilBERT, RoBERTa; Image Captioning— InceptionV3; Speech Recognition— Wav2Vec2 XLS-R & Yes. 363 languages across 32 language families & Yes \\
\hline
~\citet{ravichander-etal-2022-condaqa} CONDAQA: A Contrastive Reading Comprehension Dataset for Reasoning about Negation & 14,182 question-answer pairs & Reading Comprehension, Reasoning & Full fine-tuning (BERT, RoBERTa, DeBERTa, T5 variant namely UnifiedQA-v2-3b) AND Few-shot setting (UnifiedQA-v2-{Base, Large, 3B} , InstructGPT-orig ie. text-davinci-002) and Human Evaluation & No & No \\
\hline
~\citet{smith-etal-2022-im} “I’m sorry to hear that”: Finding New Biases in Language Models with a Holistic Descriptor Dataset & 459,758 sentences & Bias & GPT2-large, RoBERTa, DialoGPT, BlenderBot 2.0 & No & No \\
\hline
~\citet{pouran-ben-veyseh-etal-2022-mee} MEE: A Novel Multilingual Event Extraction Dataset & 50K event mentions & Event Extraction & Pipeline approach — OneIE, FourIE; Joint approach— mBERT, XLM-RoBERTa & Yes. 8 languages (English, Spanish, Portuguese, Polish, Turkish, Hindi, Korean, and Japanese) & No \\
\hline
~\citet{mekala-etal-2022-leveraging} Leveraging QA Datasets to Improve Generative Data Augmentation & N/A & Generate synthetic data (Data Augmentation) & BERT-base, BackTranslation, PEGASUS, EDA, LAMBADA & No & No \\
\hline
~\citet{chia-etal-2022-dataset} A Dataset for Hyper-Relational Extraction and a Cube-Filling Approach & 44,372 unique facts & Relational Extraction  & BERT, Feed-forward network. Pipeline Baseline—UniRE, BERT-Tagger, DistilBERT. Generative Baseline— BART & No & No \\
\hline
~\citet{zhao-etal-2022-fine-grained} A Fine-grained Chinese Software Privacy Policy Dataset for Sequence Labeling and Regulation Compliant Identification & 11K sentences, 52K fine-grained annotations & Sequence Labeling & Hidden Markov Model (HMM), Conditional Random Field (CRF), BiLSTM, BiLSTM-CRF, BERT-BiLSTM-CRF & No (chinese) & No \\
\hline
~\citet{jeong-etal-2022-kold} KOLD: Korean Offensive Language Dataset & 40,429 comments & Offensive Language detection  & Korean BERT and RoBERTa models & No (Korean) & No \\
\hline
~\citet{mukherjee-etal-2022-ectsum} ECTSum: A New Benchmark Dataset For Bullet Point Summarization of Long Earnings Call Transcripts & 2,425 document-summary pairs & Summarization & FinBERT, T5, LexRank, PacSum , SummaRuNNer, BertSumEXT, MatchSum, BART, Pegasus, BigBird, Longformer Encoder Decoder & No  & No \\
\hline
~\citet{varshney-etal-2022-cdialog} CDialog: A Multi-turn Covid-19 Conversation Dataset for Entity-Aware Dialog Generation & 1,012 dialogs and 7,982 utterances & Entity-Aware Dialog Generation & BioBERT-BASE, GPT-2, DialogGPT-finetune, BERT, BART, Human Evaluation. & No  & No \\
\hline
~\citet{pujari-etal-2022-three} Three Real-World Datasets and Neural Computational Models for Classification Tasks in Patent Landscaping & 9,465 instances (Injection-Valve) & Patent Classification & SciBERT, Transformer-based Multi-task Model & No  & No \\
\hline
\bottomrule
\end{tabular}
\caption{\label{Table: Datasets_summary_Part_V} This table shows details for the name of the paper, the size of the dataset, the NLP topic of interest, baselines and multilingualism or multimodality status of the dataset (\texttt{Part V}). }
\end{table*}
\begin{table*}[!t]
\centering
\scriptsize
\begin{tabular}{ p{4cm}| p{1.5cm} | p{2cm} | p{3cm} | p{2cm} | p{1cm}   }
\toprule
\hline 
 Paper/ Dataset & Size (\#sentences) & NLP Topic & Baselines & Multi-lingual? & Multi-modal? \\
\midrule
{\bf Track: EMNLP 2022 continued} \\
\hline
~\citet{ye-etal-2022-zerogen} ZeroGen: Efficient Zero-shot Learning via Dataset Generation & N/A & Zero-shot Learning  & GPT2, GPT2-large, GPT2-XL, LSTM, DistilBERT & No  & No \\
\hline
~\citet{zhao-etal-2022-jddc} JDDC 2.1: A Multimodal Chinese Dialogue Dataset with Joint Tasks of Query Rewriting, Response Generation, Discourse Parsing, and Summarization & 246K dialogue sessions, 3M utterances, and 507K images & multimodal dialogue response generation task, query rewriting task, dialogue discourse parsing task, dialogue summarization & BERT, k-NN, ResNet18,  GPT-2, TransResNet & No (Chinese) & Yes \\
\hline
~\citet{eklund-forsman-2022-topic} Topic Modeling by Clustering Language Model Embeddings: Human Validation on an Industry Dataset & N/A & Topic Modeling & BERT, UMAP, HDBSCAN & No & No \\
\midrule
{\bf Track: Findings of EMNLP 2022 } \\
\hline
~\citet{qu-etal-2022-commonsense} Commonsense Knowledge Salience Evaluation with a Benchmark Dataset in E-commerce & 30K entities and relations & salience of commonsense knowledge (CSK) & BERTSAGE, KG-BERT, StAR, GenKGC, PKGC,  & No  & No \\
\hline
~\citet{zhao-etal-2022-narrasum} NarraSum: A Large-Scale Dataset for Abstractive Narrative Summarization & 122K narratives & Abstractive Summarization & RANDOM, LEAD, TEXTRANK, LEXRANK, HSG, PRESUMM; Extractive summarization — BERT, ROBERTA, LONGFORMER; Abstractive summarization — BART, T5, PEGASUS, LED & No  & No \\
\hline
~\citet{zhang-etal-2022-movieun} MovieUN: A Dataset for Movie Understanding and Narrating & TNG task— 3,253 long video clips and MCN task— 33,060 video clips & Movie Understanding and Narrating & Vanilla Transformer, OVP, CLIP,  MIL-NCE, e Arcface model & No (Chinese) & Yes \\
\hline
~\citet{liu-etal-2022-long} Long Text and Multi-Table Summarization: Dataset and Method & 21,125 annual reports from 3,794 companies & long text and multi-table summarization & LexRank, TextRank, BART, PEGASUS, T5, BigBird-PEGASUS, Longformer Encoder Decoder (LED) & No  & No \\
\hline
~\citet{sun-etal-2022-visual} Visual Named Entity Linking: A New Dataset and A Baseline & 48K images, 120K named entities & Visual Entity Linking  & ResNet, CLIP,  & No  & Yes \\
\hline
~\citet{ekram-etal-2022-banglarqa} BanglaRQA: A Benchmark Dataset for Under-resourced Bangla Language Reading Comprehension-based Question Answering with Diverse Question-Answer Types & 14,889 question-answer pairs & QA & BanglaT5, mT5, BanglaBERT, mBERT &  No (Bangla) &  No \\
\hline 
~\citet{zhou-etal-2022-towards-identifying} Towards Identifying Social Bias in Dialog Systems: Framework, Dataset, and Benchmark & 28K context-response pairs & Dialog systems & Bert-Base-Chinese, VANILLA, MIXTURE-OF EXPERTS, CDIAL-GPT, EVA, CPM & No (Chinese) & No \\
\hline
~\citet{bassignana-plank-2022-crossre} CrossRE: A Cross-Domain Dataset for Relation Extraction & 5,265 sentences and 18,608 relations & Relation Extraction & BERT & No & No \\
\hline
~\citet{ye-etal-2022-progen} ProGen: Progressive Zero-shot Dataset Generation via In-context Feedback & & Dataset Generation & LSTM, DistilBERT & ?? & ?? \\
\hline
~\citet{otmakhova-etal-2022-m3} M3: Multi-level dataset for Multi-document summarisation of Medical studies &  6365 document & Multi-document summarization & BART, BioBART, Pegasus, BigBird Pegasus, Primera & No  & No \\ 

\hline
\bottomrule
\end{tabular}
\caption{\label{Table: Datasets_summary_Part_VI} This table shows details for the name of the paper, the size of the dataset, the NLP topic of interest, baselines and multilingualism or multimodality status of the dataset (\texttt{Part VI}). }
\end{table*}
\begin{table*}[!t]
\centering
\scriptsize
\begin{tabular}{ p{4cm}| p{1.5cm} | p{1.5cm} | p{3.5cm} | p{2cm} | p{1cm}   }
\toprule
\hline 
 Paper/ Dataset & Size (\#sentences) & NLP Topic & Baselines & Multi-lingual? & Multi-modal? \\
\midrule
{\bf Track: Findings of EMNLP 2022 continued} \\
\hline
~\citet{fekih-etal-2022-humset} HumSet: Dataset of Multilingual Information Extraction and Classification for Humanitarian Crises Response & 148,621 entries and 62 different tags & Multilingual Information Extraction & Entry extraction— LEAD4, XtremeDistil, XLM-R-Base and Entry classification— FastText, XtremeDistil, XLM-R-Base  & Yes (English, French, Spanish)  & No \\ 
\hline
~\citet{bastan-etal-2022-bionli} BioNLI: Generating a Biomedical NLI Dataset Using Lexico-semantic Constraints for Adversarial Examples & 21997 instances & NLI & PubMedBERT, BioLinkBERT,  & No  & No \\
\hline
~\citet{armstrong-etal-2022-jampatoisnli} JamPatoisNLI: A Jamaican Patois Natural Language Inference Dataset & 650 examples & Patois NLI & English BERT, multilingual BERT, English RoBERTa, XLM-RoBERTa & No (Patois) & No \\
\hline 
~\citet{lai-etal-2022-multilingual-subevent} Multilingual SubEvent Relation Extraction: A Novel Dataset and Structure Induction Method & 3,591 documents & Relation Extraction & RoBERTa-base, mBERT, XLMR, StructLR, TacoLM, Joint, EventSeg, SCS  & Yes. 5 languages (English, Danish, Spanish, Turkish, and Urdu) & No \\
\hline 
~\citet{liu-etal-2022-augmenting-multi} Augmenting Multi-Turn Text-to-SQL Datasets with Self-Play & -- & Text-to-SQL & PICARD/ T5-Base, T5-Large & No & No \\
\hline
~\citet{srinivasan-choi-2022-tydip} TyDiP: A Dataset for Politeness Classification in Nine Typologically Diverse Languages & 4.5K examples & Politeness Classification  & English-RoBERTa, XLM-RoBERTa, GPT3 Davinci-002 & Yes. 9 languages (Hindi, Korean, Spanish, Tamil, French, Vietnamese, Russian, Afrikaans, and Hungarian.) &  No \\ 
\hline 
~\citet{yang-etal-2022-pcmsp} PcMSP: A Dataset for Scientific Action Graphs Extraction from Polycrystalline Materials Synthesis Procedure Text & 305 documents, 2468 sentences, 14592 entities, 13968 relations & Action Graphs Extraction & BERT-base, SciBERT, MatBERT  & No  & No \\
\hline 
~\citet{liu-etal-2022-code} Code Generation From Flowcharts with Texts: A Benchmark Dataset and An Approach & 320 flowcharts & source  code generation  & LSTM, Transformer & No  & No \\ 
\hline
~\citet{maheshwari-etal-2022-benchmark} A Benchmark and Dataset for Post-OCR text correction in Sanskrit & 218,000 sentences, & Post-OCR text correction & CopyNet, mBART, mT5, ByT5, IndicBART & No (Sanskrit) & No \\
\hline
~\citet{kryscinski-etal-2022-booksum} BOOKSUM: A Collection of Datasets for Long-form Narrative Summarization & 405 documents, & Summarization & Lead-3, CNN-LSTM Extractor, BertExt, MatchSum, BART, T5, PEGASUS & No & No \\
\hline
~\citet{liu-etal-2022-wanli} WANLI: Worker and AI Collaboration for Natural Language Inference Dataset Creation & 107,885 examples & NLI & RoBERTa-large, GPT-3 & No & No \\
\hline
~\citet{yang-etal-2022-apeach} APEACH: Attacking Pejorative Expressions with Analysis on Crowd-Generated Hate Speech Evaluation Datasets & -- & hate speech detection & KoBERT, DistilKoBERT, KoELECTRA, KcBERT, SoongsilBERT  & No (Korean) & No \\
\hline
~\citet{thapa-etal-2022-multi} A Multi-Modal Dataset for Hate Speech Detection on Social Media: Case-study of Russia-Ukraine Conflict & 5,680 text-image pairs & Hate Speech Detection & ResNet+BERT, VisualBERT, CLIP; Text unimodal—LSTM, BERT, RoBERTa, and Visual unimodal—DenseNet, ResNet, VGG-19 & No  & Yes \\
\hline
~\citet{sticha-brenner-2022-hybrid} Hybrid Knowledge Engineering Leveraging a Robust ML Framework to Produce an Assassination Dataset  & 7,457 assassination events & Socio-policial event Extraction & SVM, BERT, Longformer,  & No  & No \\
\hline
~\citet{zavarella-etal-2022-tracking, hurriyetoglu-etal-2021-multilingual} ClassBases at the CASE-2022 Multilingual Protest Event Detection Task: Multilingual Protest News Detection and Automatically Replicating Manually Created Event Datasets
COVID-related protest & event datasets from the New York Times news corpus  & event detection & document classification—XLM-RoBERTa-base, mLUKE-base and XLM-RoBERTa-large AND sentence classification— mLUKE-base AND token classification—XLM-RoBERTa-base & Yes (English, Spanish and Portuguese) & No \\
\hline
\bottomrule
\end{tabular}
\caption{\label{Table: Datasets_summary_Part_VII} This table shows details for the name of the paper, the size of the dataset, the NLP topic of interest, baselines and multilingualism or multimodality status of the dataset (\texttt{Part VII}). }
\end{table*}

\section{BLOOM Dataset}
\label{sec: BLOOM_Dataset}
Table~\ref{Table: BLOOMdatasetLanguages} shows all the languages represented in the BLOOM dataset.
\begin{table*}[!t]
\centering
\footnotesize 
\begin{tabular}{ p{15.5cm}   }
\toprule
\hline 
{\bf  \centerline{The 363 Languages in BLOOM Library}   }  \\ 
\hline
Ghotuo[aaa]; Ayta, Ambala[abc]; Dangme[ada];
Adangbe[adq]; Akeu[aeu]; Afrikaans[afr];
Aghem[agq]; Esimbi[ags]; Akha[ahk]; Arosi[aia];
Amri Karbi[ajz]; Akan[aka]; Yanesha’[ame];
Amharic[amh]; Alamblak[amp]; Amuzgo,
Guerrero[amu]; Obolo[ann]; Athpariya[aph];
Awadhi[awa]; Awa[awb]; Nahuatl, Western Durango[azn]; Awing[azo]; Tuki[bag];
Bamanankan[bam]; Bambili-Bambui[baw];
Bamun[bax]; Babanki[bbk]; Balochi, Southern[bcc]; Bamenyam[bce]; Iceve-Maci[bec]; Benabena[bef]; Bengali[ben]; Bafut[bfd]; Mmen[bfm];
Bunak[bfn]; Bangandu[bgf]; Bhojpuri[bho];
Buwal[bhs]; Bislama[bis]; Banjar[bjn]; Binumarien[bjr]; Baka[bkc]; Bakoko[bkh]; Kom[bkm];
Baikeno[bkx]; Aweer[bob]; Tibetan, Central[bod];
Bozo, Tieyaxo[boz]; Wumboko[bqm]; Braj
Bhasha[bra]; Lave[brb]; Mokpwe[bri]; Bru, Western[brv]; Akoose[bss]; Ntcham[bud]; Terei[buo];
Bafaw-Balong[bwt]; Bunu, Bu-Nao[bwx];
Tairaha[bxa]; Bukusu[bxk]; Batak[bya]; Bozo,
Jenaama[bze]; Bisu[bzi]; Kaqchikel[cak];
Kakataibo-Kashibo[cbr]; Cebuano[ceb]; Kagayanen[cgc]; Chontal, Highland Oaxaca[chd];
Dene[chp]; Cimbrian[cim]; Kurdish, Central[ckb];
Chontal, Lowland Oaxaca[clo]; Chinese, Mandarin[cmn]; Chinese, Mandarin[cmn]; Mnong,
Central[cmo]; Cree, Swampy[csw]; Gichuka[cuh];
Cuvok[cuv]; Dagbani[dag]; Fataluku[ddg]; Dedua[ded]; German, Standard[deu]; Chidigo[dig];
Zarma[dje]; Kinabatangan, Upper[dmg]; Dani,
Western[dnw]; Kadazan Dusun[dtp]; Lotud[dtr];
Dotyali[dty]; Chiduruma[dug]; Elip[ekm]; Markweeta[enb]; En[enc]; English[eng]; Ewondo[ewo];
Filipino[fil]; Fali[fli]; Fon[fon]; French[fra];
Fulfulde, Adamawa[fub]; Fulfulde, Western
Niger[fuh]; Galolen[gal]; Gadaba, Bodo[gbj];
Gavar[gou]; German, Swiss[gsw]; Wayuu[guc];
Gujarati[guj]; Ekegusii[guz]; Gawri[gwc];
Hakö[hao]; Haitian Creole[hat]; Hausa[hau];
Nya Huba[hbb]; Kamwe[hig]; Hiligaynon[hil];
Hindi[hin]; Halia[hla]; Mina[hna]; Hre[hre];
Haroi[hro]; Idaté[idt]; Ilocano[ilo]; Indonesian[ind]; Inoke-Yate[ino]; Isu[isu]; Italian[ita];
Ngomba[jgo]; Mixtec, Western Juxtlahuaca[jmx];
Japanese[jpn]; Jarai[jra]; Kalanguya[kak];
Kamba[kam]; Kannada[kan]; Kamano[kbq];
Ap Ma[kbx]; Kanuri, Manga[kby]; Kanuri,
Manga[kby]; Q’eqchi’[kek]; Kenyang[ken];
Lü[khb]; Khmer[khm]; Gikuyu[kik]; Kinyarwanda[kin]; Kyrgyz[kir]; Q’anjob’al[kjb];
Kâte[kmg]; Kurdish, Northern[kmr]; Kamasau[kms]; Kanite[kmu]; Korean[kor]; Kimaragang[kqr]; Krung[krr]; Karen, S’gaw[ksw];
Lahta[kvt]; Kwaio[kwd]; Kwakum[kwu]; Khirwar[kwx]; Koli, Wadiyari[kxp]; Kenga[kyq];
Lango[laj]; Laru[lan]; Lao[lao]; Lohorung[lbr];
Lefa[lfa]; Lugbara[lgg]; Lengo[lgr]; Lhomi[lhm];
Lahu[lhu]; Lukabaras[lkb]; Lole[llg]; Limbum[lmp]; Lamnso’[lns]; Narim[loh]; Lacid[lsi];
Lutachoni[lts]; Ganda[lug]; Lawa, Eastern[lwl];
Maithili[mai]; Malayalam[mal]; Mam[mam];
Marathi[mar]; Mandar[mdr]; Matal[mfh];
Mefele[mfj]; Mpumpong[mgg]; Mambae[mgm];
Meta’[mgo]; Malila[mgq]; Lhao Vo[mhx]; Mixtec,
Ayutla[miy]; Makasae[mkz]; Manambu[mle];
Kiwilwana[mlk]; Moloko[mlw]; Mmaala[mmu];
Naba[mne]; Mundani[mnf]; Mon[mnw];
Barí[mot]; Mamasa[mqj]; Cheke Holo[mrn]; Mandaya[mry]; Masbatenyo[msb]; Muthuvan[muv];
Marwari[mve]; Mada[mxu]; Burmese[mya];
Sénoufo, Mamara[myk]; Masaaba[myx];
Mumuye[mzm]; Naasioi[nas]; Sibe[nco];
Newar[new]; Ngemba[nge]; Ngwo[ngn]; Nahuatl, Isthmus-Mecayapan[nhx]; Njyem[njy];
Ngombale[nla]; Dutch[nld]; Nahuatl, Orizaba[nlv]; Thai, Northern[nod]; Nepali[npi];
Naskapi[nsk]; Nehan[nsn]; Sotho, Northern[nso]; Naga, Tangshang[nst]; Nyole[nuj];
Ngwe[nwe]; Tanna, Southwest[nwi]; Nauete[nxa];
Nuaulu, South[nxl]; Chichewa[nya]; Nyoro[nyo]; Nyungwe[nyu]; Mbembe, Tigon[nza];
Oadki[odk]; Oji-Cree[ojs]; Okiek[oki]; Tairora,
South[omw]; Odia[ory]; Koonzime[ozm]; Pagibete[pae]; Pangasinan[pag]; Punjabi, Eastern[pan];
Pashto, Southern[pbt]; Palaung, Ruching[pce];
Paniya[pcg]; Kayan[pdu]; Indonesian, Peranakan[pea]; Persian, Iranian[pes]; Petats[pex];
Pijin[pis]; Kipfokomo[pkb]; Pamona[pmf];
Pana[pnz]; Portuguese[por]; Gapapaiwa[pwg];
Quechua, Huallaga[qub]; K’iche’[quc]; Quechua,
Lambayeque[quf]; Quechua, Cusco[quz];
Quechua, Eastern Apurímac[qve]; Quechua,
Huamalíes-Dos de Mayo Huánuco[qvh]; Quechua,
Margos-Yarowilca-Lauricocha[qvm]; Quichua,
Napo[qvo]; Quechua, Panao[qxh]; Rendille[rel];
Ranglong[rnl]; Romanian[ron]; Rotokas[roo];
Rusyn[rue]; Roviana[rug]; Russian[rus]; Sanskrit[san]; Samburu[saq]; Santhali[sat]; Sos
Kundi[sdk]; Semai[sea]; Surigaonon[sgd];
Shan[shn]; Sama, Central[sml]; Soninke[snk];
Sangil[snl]; Somali[som]; Sotho, Southern[sot];
Swo[sox]; Spanish[spa]; Saposa[sps]; Waata[ssn];
Arammba[stk]; Swahili[swh]; Swahili[swh];
Suba[sxb]; Syuba[syw]; Tamang, Eastern[taj];
Tamil[tam]; Tiang[tbj]; Panchpargania[tdb];
Emberá-Tadó[tdc]; Tamang, Western[tdg]; Tetun
Dili[tdt]; Teso[teo]; Tetun[tet]; Tajik[tgk];
Tagalog[tgl]; Thai[tha]; Tharu, Madhya
Ksetriya[the]; Kitharaka[thk]; Tharu, Dangaura[thl]; Tha[thy]; Teop[tio]; Tukudede[tkd];
Lenakel[tnl]; Tanna, North[tnn]; Whitesands[tnp];
Tontemboan[tnt]; Toma[tod]; Tombulu[tom];
Tok Pisin[tpi]; Me’phaa, Tlacoapa[tpl];
Tampuan[tpu]; Tsamai[tsb]; Setswana[tsn];
Tsonga[tso]; Turkana[tuv]; Turka[tuz]; Taveta[tvs];
Tz’utujil[tzj]; Muduga[udg]; Mundari[unr];
Urdu[urd]; Uzbek, Northern[uzn]; Venda[ven];
Vietnamese[vie]; Vili[vif]; Waray-Waray[war];
Wa, Vo[wbm]; Wagdi[wbr]; Wambon[wms];
Comorian, Ndzwani[wni]; Wanukaka[wnk];
Watakataui[wtk]; Xhosa[xho]; Kagoro[xkg];
Mbudum[xmd]; Mengaka[xmg]; Malay, Manado[xmm]; Soga[xog]; Mixtec, Yoloxóchitl[xty];
Nugunu[yas]; Yangben[yav]; Yemba[ybb];
Yakkha[ybh]; Yamphu[ybi]; Yiddish, Eastern[ydd];
Yiddish, Eastern[ydd]; Ravula[yea]; Riang
Lai[yin]; Yamap[ymp]; Zapotec, Mitla[zaw];
Malay[zlm]; Tokano[zuh]; Zulu[zul] \\
\hline 
\bottomrule
\end{tabular}
\caption{\label{Table: BLOOMdatasetLanguages}  The full list of languages in the BLOOM~\citep{leong-etal-2022-bloom} dataset.}
\end{table*}

\end{document}